\documentclass{article}
\usepackage{spconf,amsmath,graphicx}
\usepackage{graphicx}
\usepackage{xfrac}
\usepackage{wrapfig}
\usepackage{multirow}
\usepackage{booktabs}       
\usepackage{color}
\usepackage{xcolor}
\usepackage{tabularx}
\usepackage{hyperref}
\usepackage{spconf,amsmath,graphicx}

\title{Freq-Mip-AA : Frequency Mip Representation for Anti-Aliasing Neural Radiance Fields}
\name{Youngin Park$^{1,4}$ 
    \qquad Seungtae Nam$^{2}$ 
    \qquad Cheul-hee Hahm$^{4}$ 
    \qquad Eunbyung Park$^{2,3,*}$\thanks{* Corresponding author.}}
    \address{$^{1}$ Department of Digital Media and Communications Engineering, Sungkyunkwan University \\
      $^{2}$ Department of Artificial Intelligence, Sungkyunkwan University \\
      $^{3}$ Department of Electrical and Computer Engineering, Sungkyunkwan University \\
      $^{4}$ Visual Display Division, Samsung Electronics}

\begin{document}
%
\maketitle
\begin{abstract}
Neural Radiance Fields (NeRF) have shown remarkable success in representing 3D scenes and generating novel views. 
However, they often struggle with aliasing artifacts, especially when rendering images from different camera distances from the training views. 
To address the issue, Mip-NeRF proposed using volumetric frustums to render a pixel and suggested integrated positional encoding (IPE). While effective, this approach requires long training times due to its reliance on MLP architecture. In this work, we propose a novel anti-aliasing technique that utilizes grid-based representations, usually showing significantly faster training time. In addition, we exploit frequency-domain representation to handle the aliasing problem inspired by the sampling theorem. The proposed method, FreqMipAA, utilizes scale-specific low-pass filtering (LPF) and learnable frequency masks. Scale-specific low-pass filters (LPF) prevent aliasing and prioritize important image details, and learnable masks effectively remove problematic high-frequency elements while retaining essential information. By employing a scale-specific LPF and trainable masks, FreqMipAA can effectively eliminate the aliasing factor while retaining important details. We validated the proposed technique by incorporating it into a widely used grid-based method. The experimental results have shown that the FreqMipAA effectively resolved the aliasing issues and achieved state-of-the-art results in the multi-scale Blender dataset. Our code is available at \href{https://github.com/yi0109/FreqMipAA}{https://github.com/yi0109/FreqMipAA}.
\end{abstract}

\section{Introduction}
\label{sec:intro}

Neural Radiance Fields (NeRF)~\cite{nerf} have been remarkably successful in representing 3D scenes and generating novel views.
However, assuming cameras are located at the same distance from the center of the scene, NeRF often suffers from aliasing artifacts (e.g., jaggedness and blurriness) when rendering images at varying camera distances.
This assumption oversimplifies real-world scenarios, where objects are often located at varying distances from the cameras.
Mip-NeRF~\cite{mip-nerf} has addressed this issue by rendering the conical frustums instead of rays, but their architecture heavily relies on a large MLP and requires a significantly long training time.


Recently proposed grid-based representations greatly accelerate the training time of NeRF by directly optimizing the positional information into learnable feature vectors that can be decoded into color and density values with only a tiny MLP, but the aliasing issue still remains.
A line of works built alias-free NeRF based on grid representations, mitigating the slow training time of mip-NeRF.
They apply fixed average kernels~\cite{hu2023tri} or learnable convolution filters~\cite{nam2023mip} to a single-scale grid and generate multi-scale grids, where each of them can represent different scales of a scene.
Though effective, the discussions are limited to the spatial domain, and a direct application of these works to the frequency domain is not straightforward.
Since handling the aliasing issue in the frequency domain is much easier than in the spatial domain, where one can simply limit the maximum frequency to the Nyquist rate of a given signal, it is important to explore the frequency perspective of the problem.

Inspired by the conventional image analysis techniques in the frequency domain~\cite{schonewille2009anti, gonzalez2009digital, buades2005review, burt1987laplacian}, we propose a novel approach that directly handles the aliasing issue of the grid-based NeRFs in the frequency domain.
The proposed method directly optimizes a single-scale grid representation and multiple frequency masks in the frequency domain, and multi-scale grid representations are generated by applying element-wise multiplication between the single-scale grid and the frequency masks followed by the inverse Discrete Cosine Transform (DCT).
Each frequency mask limits the maximum frequency of the single-scale grid by simply removing the high-frequency components that are unnecessary for representing a requested signal.
Furthermore, when training the single-scale grid, we apply a series of fixed Gaussian low-pass filters with increasing standard deviations to make the model learn the most important low-frequency components first and progressively add detailed information about the high-frequency components.

To test the effectiveness of our approach, we integrated FreqMipAA into one of the recent grid-based NeRFs: TensoRF~\cite{tensorf}.
We trained and tested our model on the multi-scale Blender Dataset, and experimental results showed that our model achieves the best average PNSR (Table~\ref{table:main_result}), outperforming the prior art, Tri-MipRF~\cite{hu2023tri}.
We also conducted an ablation study to demonstrate that the learnable frequency masks are important to remove the aliasing artifacts (e.g., jaggedness) in low-resolution images.
The contributions of this paper can be summarized as follows:
\begin{itemize}
     \item To the best of our knowledge, it is the first attempt to exploit frequency domain representations to resolve aliasing issues in radiance fields.
     
     \item We introduce a novel training method that effectively mitigates aliasing in NeRF by employing learnable frequency masks and scale-specific low-pass filters.

     \item The proposed method achieves a state-of-the-art Peak Signal-to-Noise Ratio (PSNR) on the multi-scale Blender dataset, showcasing its better rendering quality.
     
 \end{itemize}

\section{Related works}
\label{sec:related}

\subsection{Anti-aliasing}
\label{ssec:anti-aliasing}

The issue of anti-aliasing is a fundamental barrier in computer graphics and neural volumetric rendering. It requires advanced techniques to reduce the visual distortions caused by overlapping frequency components. Conventional methods, such as supersampling~\cite{barron2023zipnerf} and prefiltering~\cite{turki2023pynerf}, have established the foundation for resolving this problem. Supersampling techniques enhance the sampling rate to approximate the Nyquist frequency, substantially decreasing aliasing. However, they require significant processing resources, making them better suited for offline rendering scenarios. On the other hand, mipmap~\cite{williams1983pyramidal}, which is one of the prefiltering techniques~\cite{mip-nerf,Mip-NeRF_360}, provides effective solutions for rendering in real-time. It achieves this by reducing the Nyquist frequency requirement and adjusting the level of detail according to the viewer's distance. As a result, it avoids the high computational load that comes with supersampling. Building upon these fundamental principles, we provide a technique that focuses on efficiently decreasing the aliasing factor through frequency domain filtering. 

\subsection{Grid-based Model}

To address the significant storage efficiency achieved by NeRF's~\cite{nerf} implicit scene representation through an MLP, its notable drawback lies in the prolonged rendering and reconstruction times. In pursuit of accelerating NeRF's training process, extensive investigations have ventured into grid-based representations, significantly reducing training durations to under an hour while maintaining rendering efficacy. The evolution within grid-based frameworks has seen the introduction of diverse strategies aimed at bolstering convergence rates, including the deployment of voxel grids~\cite{sun2022direct}, the application of low-rank tensors~\cite{tensorf}, and the innovation of hash tables~\cite{instant-ngp,wang2023f2nerf}, each contributing to the precise capture of spatial attributes crucial for the detailed rendering of color and density.
Further efforts to merge NeRF's implicit format with voxel octrees~\cite{liu2020neural} have been made to quicken inference speeds. Notably, Plenoxels~\cite{Plenoxels} achieves a leap in efficiency by optimizing voxel grids directly as opposed to MLPs, thus facilitating training durations that span mere minutes or seconds. Similarly, TensoRF~\cite{tensorf} and K-Planes~\cite{kplanes_2023} introduce low-rank tensor approximations to voxel grids, effectively minimizing memory demands. Building upon previous studies, FreqMipAA aims to eliminate aliasing by performing effective adjustments in grid representations.

\subsection{Anti-aliasing in NeRF}
Several studies have presented methods to address anti-aliasing in NeRF. Mip-NeRF uses a 3D conical frustum to address aliasing problems. Building on InstantNGP's hash grid representation, Zip-NeRF~\cite{barron2023zipnerf} proposes a multi-sampling technique in the conical frustum rather than the camera ray., but does not include position encoding.  Tri-MipRF~\cite{hu2023tri} introduces tri-plane mipmaps to represent multi-scale 3D scenes. MipGrid~\cite{nam2023mip} introduced a straightforward convolutional approach to capture many scales. This structure infers lower-frequency representations by progressive down sampling. While prior research~\cite{nam2023mip,hu2023tri,turki2023pynerf} has mostly concentrated on resolving the problem of aliasing in the spatial domain, our method aims to address this issue by exerting more precise control over the frequency domain responsible for generating aliasing. We present an innovative approach to address the issue of aliasing in the frequency domain.

\section{Preliminaries}
\label{sec:prelminaries}

\subsection{Sampling Theorem}

\begin{figure*}[ht]
  \centering
   \includegraphics[width=1\linewidth]{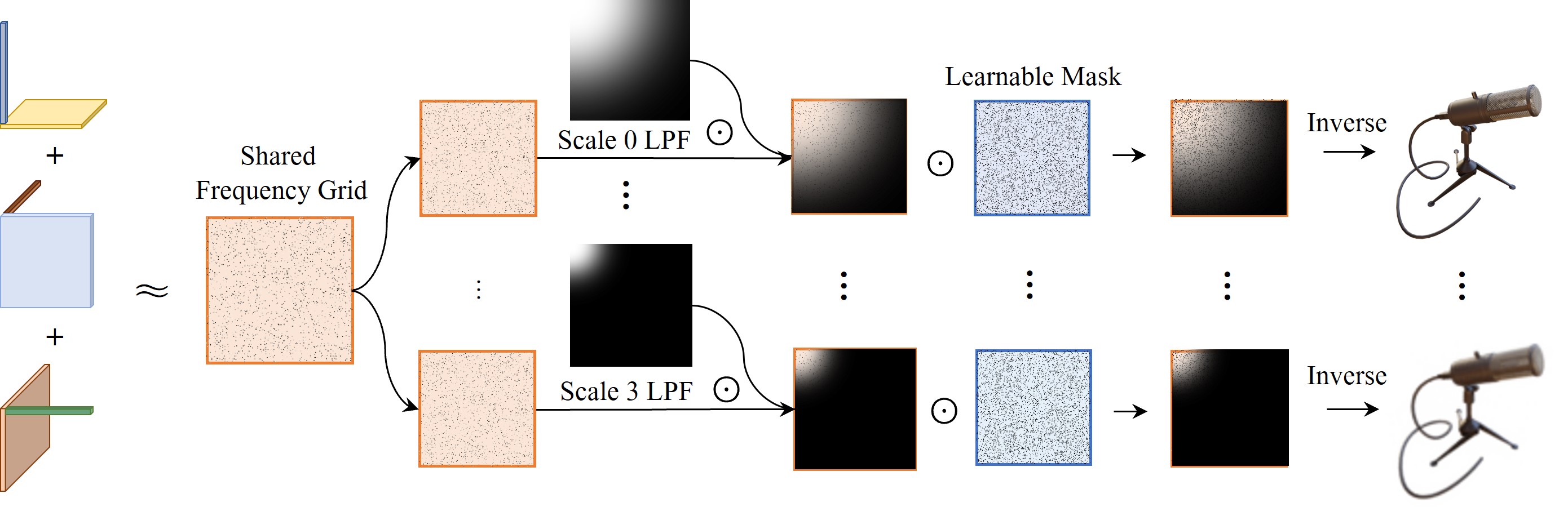}
   \caption{The overall architecture of our model begins with training a shared grid in the frequency domain. This is followed by scale-specific low-pass filters designed to facilitate focus on crucial information. Subsequently, learnable frequency masks are applied to further refine frequency grids. To enhance visual clarity, the grid is shown as a basic square shape, even though it is fundamentally a vector matrix structure. The $\odot$ represents element-wise multiplication.}
   \label{fig:overview}
\end{figure*}

The Sampling Theorem, also known as the Nyquist-Shannon Theorem, is a fundamental principle in signal processing that establishes the minimum sampling rate necessary for the accurate reconstruction of continuous signals from their samples. This theory, formulated by Claude Shannon, asserts that a continuous signal, limited to a maximum frequency of \(f_{max}\) hertz, can be fully reconstructed without any loss of information if it is sampled at a rate greater than \(2f_{max}\) samples per second, known as the Nyquist rate. This critical requirement is mathematically expressed as \(f_s > 2f_{max}\), where \(f_s\) denotes the sampling frequency—the rate at which samples are taken from the continuous signal—and \(f_{max}\) represents the signal's highest frequency component. Ensuring that the sampling rate meets or exceeds the Nyquist rate is essential for accurately recreating the original continuous signal. Conversely, sampling below this rate can introduce aliasing artifacts, compromising the integrity of the reconstructed signal.

\subsection{NeRF}
Neural Radiance Fields (NeRF) aim to accurately describe a scene from different perspectives by estimating color (\(c\)) and density (\(\sigma\)) along the rays that traverse the image. NeRF receives a position vector \(\mathbf{x} = (x, y, z)\) and a viewing direction vector \(\mathbf{d} = (\theta, \phi)\) as input, and produces the corresponding color and density values at that point. The expression \(F_{\Theta}(\mathbf{x}, \mathbf{d})\) can be represented as \((c, \sigma)\). The rendering equation in integral form, which NeRF uses to determine the color \(C\) of a ray \(r\) as it passes across the scene, is as follows:
\begin{equation}
\begin{aligned}
C(r) &= \int_{t_n}^{t_f} T(t) \sigma(r(t)) c(r(t), d) dt, \\
\text{where} \quad T(t) &= \exp\left(-\int_{t_n}^{t} \sigma(r(s)) ds\right).
\end{aligned}
\label{eq:nerf_render}
\end{equation}
The ray equation \(r(t) = o + td\) uses \(o\) as the origin, \(d\) as the direction, and \(t_n\) and \(t_f\) as near and far integration bounds, respectively. \(T(t)\) represents the accumulated transmittance to point \(t\), which is the amount of light that reaches the point without being absorbed by it. This approach replicates scene light transfer for photorealistic NeRF representations. The function \(F_{\Theta}\), often a neural network, predicts color and density from sparse input pictures, enabling complex 3D reconstruction and new perspectives.

\subsection{Grid-based NeRF}
Grid-based Neural Radiance Fields (NeRF) models, like TensoRF, make notable progress in representing volumetric scenes by utilizing efficient, high-dimensional tensor decompositions. TensoRF utilizes a vector-matrix (VM) decomposition technique to encode scene geometry and appearance more effectively, requiring fewer parameters compared to conventional NeRF models. VM decomposition is a process that breaks down a tensor into several vectors and matrices. The factorization of the 3D geometry grid $\mathcal{G}$ is expressed as:
\begin{equation}
\mathcal{G} = \sum_{r=1}^{R} v_{r}^X \circ M_{\sigma, r}^{YZ} + v_{r}^Y \circ M_{\sigma, r}^{XZ} + v_{r}^Z \circ M_{\sigma, r}^{XY}.
\end{equation}
\label{eq:tensoRF_grid}

\section{Method}
\label{sec:method}
Our technique diverges from conventional methods that address aliasing within the spatial domain. Alternatively, we implement a new approach that addresses high-frequency components that result in aliasing by utilizing a frequency domain method. The Discrete Cosine Transform (DCT) is employed as a transformation technique because of its capacity to significantly and proficiently increase the training process, leading to enhanced performance. As illustrated in Figure \ref{fig:overview}, our approach entails training a shared grid in the frequency domain and subsequently duplicating it across multiple scales. This step is followed by the application of a low-pass filter across various scales, enabling the refinement of the frequency grid. Learnable frequency masks are subsequently applied, enhancing our model's ability to selectively process frequency components. The features captured in this manner are then inversely transformed back to their original spatial domain using the inverse DCT, facilitating visualization and further processing. This process underscores our system's capacity to handle high-frequency data with precision, significantly reducing aliasing effects while maintaining computational efficiency.

\subsection{Scale-specific Low-pass filter}
We employed a Gaussian top-left filter to generate a low-pass filter. 
The Nyquist-Shannon sampling theorem establishes the correlation between sample rate and the highest frequency that can be reliably depicted. According to the theorem, reducing an image's resolution by a factor of 2 lowers the maximum accurately depicted frequency by half. This decrease implies that frequencies above the new Nyquist limit must be suppressed to avoid aliasing in the downsampled image. To reduce signal strength, a Gaussian low-pass filter with an appropriate standard deviation (\(\sigma\)) can be employed for the desired cutoff frequency.
However, calculating the value of \(\sigma\) for a Gaussian mask in the DCT domain is challenging due to the fact that the discrete cosine transform (DCT) turns an image into a set of discrete frequency components. 
Assuming a uniform distribution of Discrete Cosine Transform (DCT) coefficients that reflect frequencies ranging from 0 to the Nyquist limit, the initial value of \(\sigma\) can be approximated using the desired reduction factor \(n\) for resolution.
\begin{equation}
\sigma = \frac{N}{2n}
\end{equation}
\(N\) denotes the dimension of the Discrete Cosine Transform (DCT) domain. The purpose of this calculation of \(\sigma\) is to reduce the amplitudes of frequencies that are higher than the modified midpoint of the DCT spectrum, which corresponds to the new effective Nyquist frequency of the downsampled resolution.
 By discretely calculating and applying \(\sigma\) for each scale, we ensure that our low-pass filtering is adjusted to the resolution reduction factor (\(n\)) at each scale.
Calculating the precise value of \(\sigma\) for the Gaussian low-pass filter (\(\sigma = \frac{N}{2n}\)) is an important method for reducing aliasing by suppressing high-frequency components beyond the Nyquist limit. However, finding the optimal \(\sigma\) for different applications at different scales is still difficult. The difficulty stems from the discrete character of data in the DCT domain and the various impacts of downsampling on aliasing at different sizes. 

Through executing this procedure, we acquire the refined characteristics, \(F_\text{filtered}\), which are essential for the upcoming stages of our approach:
\begin{equation}
F_\text{filtered} = \text{LPF}(F_\text{grid}, \sigma)
\end{equation}
In this context, \(F_\text{grid}\) refers to the original grid features in the frequency domain, while \(\text{LPF}(\cdot, \sigma)\) represents the low-pass filtering operation that is performed using the estimated standard deviation (\(\sigma\)). 
In order to address the difficulty of determining the optimal \(\sigma\) for our Gaussian low-pass filters, we improve our methodology by introducing trainable masks in the subsequent section. This approach improves the ability of our model to minimize aliasing while also guiding its learning toward important information at different scales.  

\begin{table*}[t]
\centering
\scriptsize
\caption{Quantitative analysis: A detailed comparison of FreqMipAA with other models on the multi-scale blender dataset~\cite{mip-nerf}}
\vspace{0.5em}
\begin{tabular}{@{\hskip 0.55em}l@{\hskip 0.55em}|@{\hskip 0.55em}c@{\hskip 0.35em}c@{\hskip 0.55em}c@{\hskip 0.55em}c@{\hskip 0.55em}@{\hskip 0.55em}c|@{\hskip 0.55em}c@{\hskip 0.55em}c@{\hskip 0.55em}c@{\hskip 0.55em}@{\hskip 0.55em}c@{\hskip 0.35em}c|@{\hskip 0.55em}c@{\hskip 0.55em}c@{\hskip 0.55em}@{\hskip 0.55em}c@{\hskip 0.35em}c@{\hskip 0.55em}c@{\hskip 0.55em}c@{\hskip 0.55em}}
    \toprule
     & \multicolumn{5}{c|@{\hskip 0.55em}}{PSNR$\uparrow$} & \multicolumn{5}{c|@{\hskip 0.55em}}{SSIM$\uparrow$} & \multicolumn{5}{c@{\hskip 0.55em}}{LPIPS$\downarrow$} \\
     & Full Res. & \sfrac{1}{2}\:Res. & \sfrac{1}{4}\:Res. & \sfrac{1}{8}\:Res. & Avg &  Full Res. & \sfrac{1/2}\:Res. & \sfrac{1/4}\:Res. & \sfrac{1}{8}\:Res. & Avg. & Full Res. & \sfrac{1}{2}\:Res. & \sfrac{1}{4}\:Res. & \sfrac{1}{8}\:Res. & Avg. \\
    \midrule
    NeRF~\cite{nerf} & 29.90 & 32.13 & 33.40 & 29.47 & 31.23 & 0.938 & 0.959 & 0.973 & 0.962 & 0.958 & 0.074 & 0.040 & 0.024 & 0.039 & 0.044 \\
    MipNeRF~\cite{mip-nerf} & 32.63 & 34.34 & 35.47 & 35.60 & 34.51 & 0.958 & 0.970 & 0.979 & 0.983 & 0.973 & 0.047 & 0.026 & 0.017 & 0.012 & 0.026 \\
    Plenoxels~\cite{Plenoxels} & 31.60 & 32.85 & 30.26 & 26.63 & 30.34 & 0.956 & 0.967 & 0.961 & 0.936 & 0.955 & 0.052 & 0.032 & 0.045 & 0.077 & 0.051 \\
    TensoRF~\cite{tensorf} & 32.11 & 33.03 & 30.45 & 26.80 & 30.60 & 0.956 & 0.966 & 0.962 & 0.939 & 0.956 & 0.056 & 0.038 & 0.047 & 0.076 & 0.054 \\
    Instant-ngp~\cite{instant-ngp} & 30.00 & 32.15 & 33.31 & 29.35 & 31.20 & 0.939 & 0.961 & 0.974 & 0.963 & 0.959 & 0.079 & 0.043 & 0.026 & 0.04 & 0.047 \\
    Tri-MipRF~\cite{hu2023tri} & \textbf{33.32} & 35.02 & 35.78 & 36.13 & 35.06 & \textbf{0.961} & 0.974 & 0.981 & \textbf{0.986} & \textbf{0.976} & \textbf{0.043} & \textbf{0.024} & \textbf{0.017} & \textbf{0.011} & \textbf{0.024} \\
    \midrule
    FreqMipAA (Ours) & 32.49 & \textbf{35.36} & \textbf{36.70} & \textbf{36.92} & \textbf{35.37} & 0.959 & \textbf{0.976} & \textbf{0.983} & \textbf{0.986} & \textbf{0.976} & 0.054 & 0.028 & 0.020 & 0.016 & 0.029 \\
    \bottomrule
\end{tabular}
\label{table:main_result}
\end{table*}

\begin{table*}[t]
\centering
\scriptsize
\caption{Results of the ablation Study: Assessing the effect of learnable Masks and scale-specific low-pass filtering on anti-aliasing performance in FreqMipAA on the multi-scale blender dataset~\cite{mip-nerf}}
\vspace{0.5em}
\begin{tabular}{@{\hskip 0.55em}l@{\hskip 0.55em}|@{\hskip 0.55em}c@{\hskip 0.35em}c@{\hskip 0.55em}c@{\hskip 0.55em}c@{\hskip 0.55em}@{\hskip 0.55em}c|@{\hskip 0.55em}c@{\hskip 0.55em}c@{\hskip 0.55em}c@{\hskip 0.55em}@{\hskip 0.55em}c@{\hskip 0.35em}c|@{\hskip 0.55em}c@{\hskip 0.55em}c@{\hskip 0.55em}@{\hskip 0.55em}c@{\hskip 0.35em}c@{\hskip 0.55em}c@{\hskip 0.55em}c@{\hskip 0.55em}}
    \toprule
     & \multicolumn{5}{c|@{\hskip 0.55em}}{PSNR$\uparrow$} & \multicolumn{5}{c|@{\hskip 0.55em}}{SSIM$\uparrow$} & \multicolumn{5}{c@{\hskip 0.55em}}{LPIPS$\downarrow$} \\
     & Full Res. & \sfrac{1}{2}\:Res. & \sfrac{1}{4}\:Res. & \sfrac{1}{8}\:Res. & Avg. &  Full Res. & \sfrac{1}{2}\:Res. & \sfrac{1}{4}\:Res. & \sfrac{1}{8}\:Res. & Avg. & Full Res. & \sfrac{1}{2}\:Res. & \sfrac{1}{4}\:Res. & \sfrac{1}{8}\:Res. & Avg. \\
    \midrule
    FreqMipAA (w/o Learnable Mask) & 31.32 & 33.96 & 34.87 & 35.74 & 33.96 & 0.951 & 0.970 & 0.978 & 0.983 & 0.970 & 0.067 & 0.037 & 0.027 & 0.022 & 0.038 \\
    FreqMipAA (w/o LPF) & 32.19 & 34.98 & 35.76 & 33.87 & 34.20 & 0.956 & 0.973 & 0.980 & 0.978 & 0.972 & 0.059 & 0.033 & 0.026 & 0.032 & 0.038 \\
    FreqMipAA (w/o scale-specific) & 32.37 & 35.18 & 36.27 & 35.5 & 34.80 & 0.956 & 0.974 & 0.980 & 0.977 & 0.972 & 0.059 & 0.033 & 0.026 & 0.032 & 0.037 \\
    \midrule
    FreqMipAA & 32.49 & 35.36 & 36.70 & 36.92 & 35.37 & 0.959 & 0.976 & 0.983 & 0.986 & 0.976 & 0.054 & 0.028 & 0.020 & 0.016 & 0.029 \\
    \bottomrule
\end{tabular}
\label{table:abliation_result}
\end{table*}

\subsection{Learnable Frequency Mask}
\begin{figure}[t]
  \centering
   \includegraphics[width=1\linewidth]{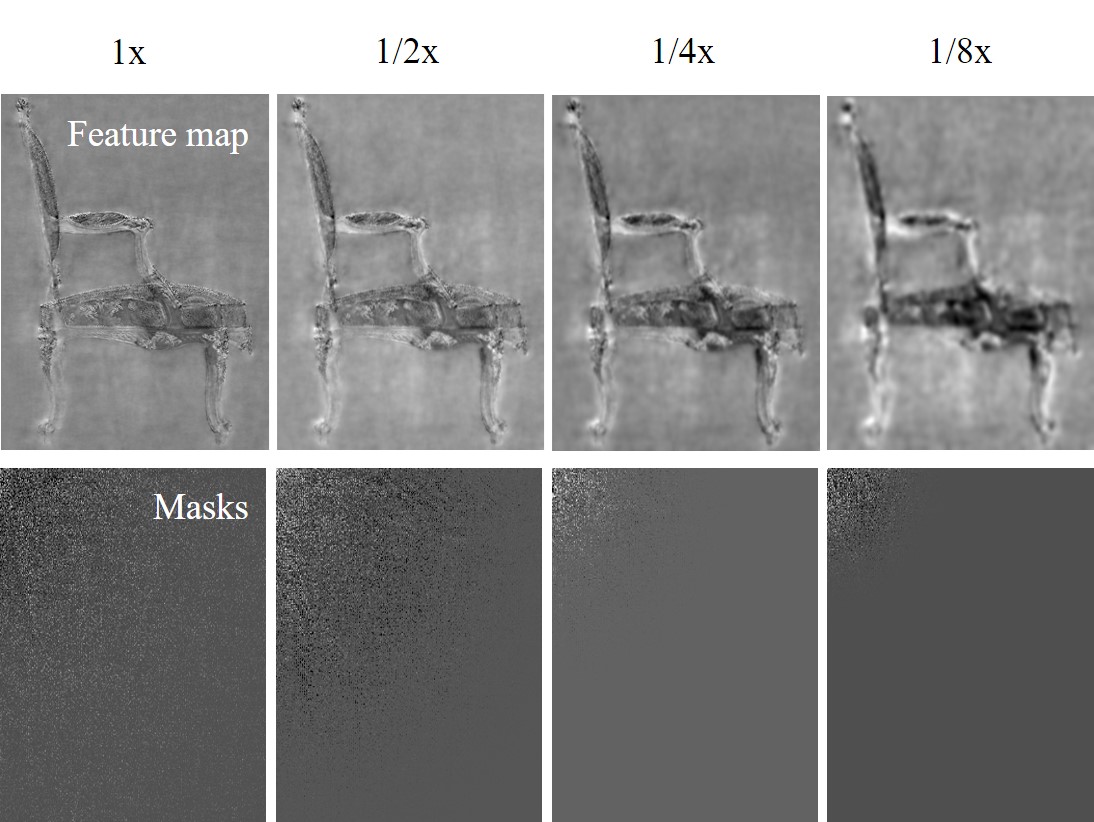}
   \caption{Scale-specific feature maps(Top) and masks(Bottom) at 1x, 1/2x, 1/4x, and 1/8x capture global structure and local details for improved picture reconstruction}\label{fig:feature_mask}
\end{figure}

Following the implementation of a low-pass filter on the grid features at each scale, we use a trainable mask to further enhance the processing of the filtered information. By implementing this novel approach, our model gains the capability to effectively differentiate between important and less important characteristics in the filtered spectrum. This significantly improves the model's capacity to prioritize the preservation of relevant details, hence boosting the quality of picture reconstruction. By incorporating a trainable mask at different scales, the filtering process becomes adaptive. This means that the model can learn to give importance to certain frequency components or suppress them, depending on their significance to the overall image quality and content representation.
In order to put this principle into practice, we implement the learnable mask onto the grid features in the following manner:
\begin{equation}
F_\text{masked} = \text{sigmoid}(F_\text{filtered} \odot M) - \epsilon
\end{equation}
The variables in the equation are \(M\), which is a mask that may be adjusted through learning, and \(\odot\), which denotes element-wise multiplication. The sigmoid activation function constrains the output \(F_\text{masked}\) to a bounded range. The activation level is adjusted precisely by utilizing a threshold $\epsilon$ to suppress characteristics that fall below a designated degree of importance. 
The masked features \(F_\text{masked}\) are subsequently restored to their original spatial domain using inverse Discrete Cosine Transform (IDCT).
The feature maps and masks representing each learned scale can be seen in Figure \ref{fig:feature_mask}.

\section{Experiments}
\label{sec:experiments}
\subsection{Experiment setup}

\begin{figure*}[ht]
  \centering
   \includegraphics[width=1\linewidth]{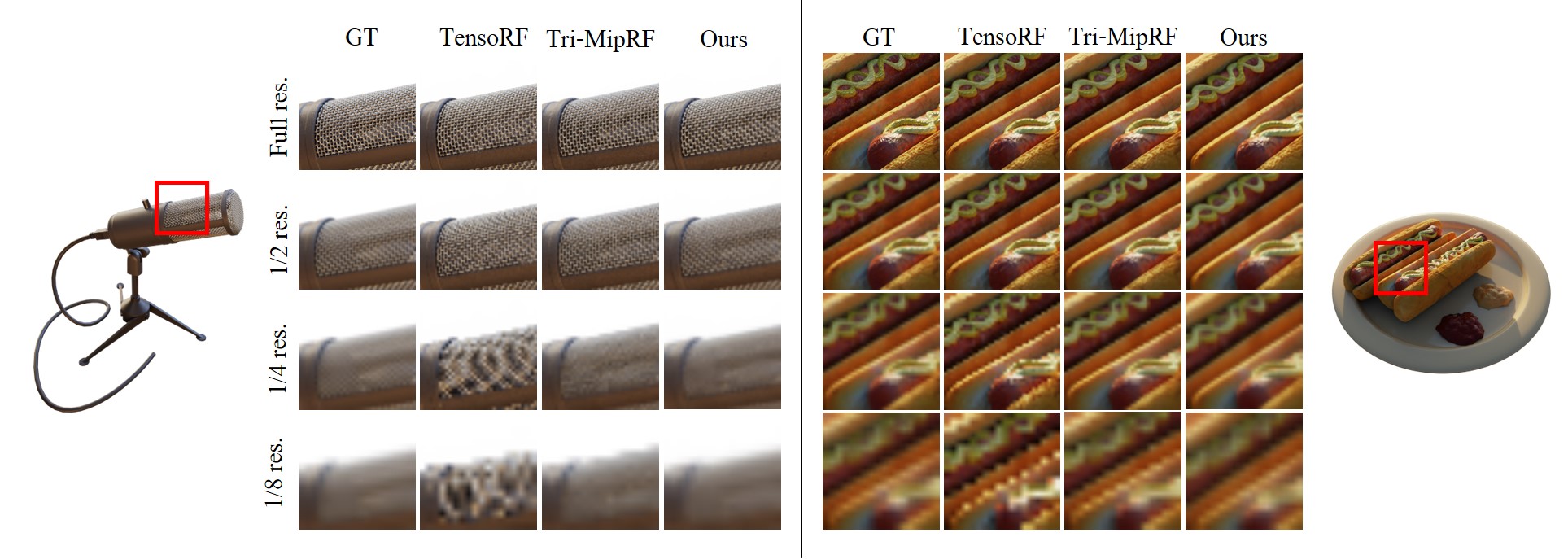}
   \caption{A qualitative evaluation of FreqMipAA against the Ground Truth, TensoRF and Tri-MipRF on Blender dataset. The zoomed sections of images rendered at four distinct scales are displayed.}
   \label{fig:exp_img}
\end{figure*}

The FreqMipAA model was derived from TensoRF. We achieved consistency by using the same TensoRF setups for all initial training hyperparameters. By employing this method, we were able to directly assess the efficacy of our frequency domain anti-aliasing technique by comparing our findings with those of the baseline TensoRF models. 
TensoRF utilizes a method of steadily improving resolution through progressive upsampling. Nevertheless, we noticed that commencing training in the frequency domain at the beginning of the upsampling procedure resulted in a substantial loss of information. In order to address this issue, we initiated our frequency domain adaptation process once the initial 7,000 steps of upsampling were completed. By strategically scheduling the process of upsampling, we were able to maintain the accuracy of the process and take advantage of the advantages of frequency domain learning for reducing aliasing. Following that, we followed the FreqMipAA process as previously described, expanding our training to a total of 40,000 iterations. We selected the Discrete Cosine Transform (DCT) as our transformation method due to its temporal and computational efficiency, as mentioned earlier. We utilize the PyTorch framework to construct our FreqMipAA, without using the tiny-cuda-nn extension. Future efforts will enhance the efficiency of FreqMipAA by utilizing the tiny-cuda framework.

\subsection{Results}
The evaluation we conducted, presented in Table \ref{table:main_result}, clearly shows how FreqMipAA performs in comparison to other advanced models on the Blender dataset. Using commonly accepted evaluation metrics in previous studies, such as Peak Signal-to-Noise Ratio (PSNR), Structural Similarity Index Measure (SSIM)~\cite{wang2004image}, and Learned Perceptual Image Patch Similarity (LPIPS)~\cite{zhang2018unreasonable} with a VGG backbone, our results demonstrate that FreqMipAA outperforms others in terms of PSNR and SSIM. This demonstrates a notable enhancement in the clarity and strength of the image, highlighting the success of our method in maintaining precise and accurate information.
Notably, FreqMipAA's performance advantage becomes even more pronounced as the resolution decreases, highlighting its exceptional capability in mitigating aliasing effects in lower resolution scenarios. This outcome is particularly relevant for Neural Radiance Fields (NeRF) applications, where maintaining visual quality across varying resolutions is a critical challenge.
Figure \ref{fig:exp_img} depicts the comparison of results at different resolutions among Ground Truth, TensoRF, and our research. This comparison showcases the efficacy of our approach in addressing the aliasing problems that are in the baseline models.
The integration of FreqMipAA with TensoRF, requiring minimal modifications, has not only facilitated a straightforward implementation but also ensured efficient training times. 

\subsection{Ablation studies}
Table \ref{table:abliation_result} presents an ablation analysis that verifies the accuracy of our claims about the significance of both learnable masks and low-pass filtering (LPF) in enhancing the anti-aliasing capabilities of the FreqMipAA model. As explained in the technique, the low-pass filter is crucial in guiding the model's attention towards important frequency components, thereby enabling a more precise approach to reducing aliasing. The absence of a low-pass filter (LPF) is most evident, especially at lower resolutions. Without the LPF, high-frequency artifacts are not effectively suppressed, leading to a decrease in image quality. As explained in the technique, the low-pass filter is crucial in guiding the model's attention towards important frequency components, thereby enabling a more precise approach to reducing aliasing. This is clearly noticeable in the decline in performance that is observed when the low-pass filter (LPF) is not present, particularly at lower resolutions.
Conversely, the absence of a learnable mask primarily affects performance at higher resolutions, where its impact becomes more pronounced. This result implies that masks that can be learned are essential for preserving accuracy in high-resolution outputs, where the subtle interaction of frequency components becomes more and more important. In the absence of learnable masks for adaptive refinement, the model is unable to reach the desirable equilibrium between reducing aliasing and preserving details, especially in high-resolution scenarios.
Furthermore, we are conducting experiments using the same Low Pass Filter (LPF). Selecting an optimal value for the single mask \(\sigma\) parameter to improve the entire range of scales is a challenging task. This study demonstrates that the use of scale-specific masks is more effective in representing multiple scales. 
This ablation study illustrates the essential role of the synergy between scale-specific LPF and learnable masks in the success of the FreqMipAA model. This combination not only offers a comprehensive solution to the problem of aliasing across all resolutions but also ensures consistent and excellent performance, even at higher resolutions.

\section{Conclusion}
\label{sec:conclusion}
The study introduces a new approach called FreqMipAA, which effectively addresses the issue of aliasing by employing frequency domain learning. This approach has been proven to be effective by demonstrating superiority in PSNR scores compared to other methods. Furthermore, our research indicates the possibility of creating compressed anti-aliasing models by training in the frequency domain. The prospect of compression is a promising direction for further investigation, as it implies that models can be enhanced in terms of efficiency without compromising quality. Additional investigation is required to optimize and implement these compressible models in order to fully exploit their potential and take advantage of the benefits provided by frequency domain learning for anti-aliasing solutions.

\section{Acknowledgement}
\label{sec:acknowledgement}
This research was supported in parts by the grant (RS-2023-00245342) from the Ministry of Science and ICT of Korea through the National Research Foundation (NRF) of Korea, the Institute of Information and Communication Technology Planning Evaluation grant (IITP-2019-0-00421) for the AI Graduate School program, Korea Institute of Energy Technology Evaluation and Planning (KETEP) and the Ministry of Trade, Industry \& Energy (MOTIE) of the Republic of Korea (No. 20224000000360). and the Culture, Sports, and Tourism R\&D Program through the Korea Creative Content Agency grant funded by the Ministry of Culture, Sports and Tourism in 2024 (Project Name: Research on neural watermark technology for copyright protection of generative AI 3D content, RS-2024-00348469).

\bibliographystyle{IEEEbib}
\bibliography{reference}

\noindent
\fbox{%
    \parbox{\columnwidth}{%
        \textbf{Copyright Notice: © 2024 IEEE. Personal use of this material is permitted. Permission from IEEE must be obtained for all other uses, in any current or future media, including reprinting/republishing this material for advertising or promotional purposes, creating new collective works, for resale or redistribution to servers or lists, or reuse of any copyrighted component of this work in other works.}
    }%
}

\end{document}